\documentclass[letterpaper,10pt]{article}
\usepackage{setspace}
\usepackage{amsfonts}
\usepackage{graphicx}
\usepackage[left=2cm,top=1cm,right=3cm,includehead,includefoot]{geometry}

\title{Admissibility of a posterior predictive decision rule}
\author{Giri Gopalan\footnote{gopalan88@gmail.com}}
\date{}
\begin{document}
\maketitle
\begin{abstract}
Recent decades have seen an interest in prediction problems for which Bayesian methodology has been used ubiquitously. Sampling from or approximating the posterior predictive distribution in a Bayesian model allows one to make inferential statements about potentially observable random quantities given observed data. The purpose of this note is to use statistical decision theory as a basis to justify the use of a posterior predictive distribution for making a point prediction.\end{abstract}
\section{Introduction and Motivation}
As reviewed by [Owhadi and Scovel], the field of statistical decision theory introduced by Wald, building on a game theoretic foundation developed by von Neumann and Morgenstern, provides links between Bayesian and frequentist statistical philosophies through the concepts of decision rules, admissibility, and risk functions amongst others. Moreover, a recent thrust of research motivated by machine learning has put much emphasis on prediction problems for which Bayesian methodology has been widely used. The purpose of this note is to demonstrate that classic decision theoretic results can be simply applied to the analysis of prediction problems. In fact, both [Berger] and [Robert] remark upon the ease of applying statistical decision theory within the context of prediction, however, no explicit result is stated in either work; the contribution of this note, therefore, is to highlight a simple way in which the results of statistical decision theory might apply to prediction problems. To the author's knowledge the most similar lines of thought appear in work by [Nayak and El-Baz], where the loss function depends on the underlying parameter (in contrast to what follows). Additionally, in [Nayak and El-Baz],  a proof of admissibility of a Bayesian prediction rule is not given, albeit alluded to.

In statistical decision theory it is desired that an estimator is admissible, meaning that there exists no other estimator whose (frequentist) risk is both: i) at least as small for all values of the parameter space and ii) strictly smaller for at least a single value of the parameter space. In this note, the notion of the risk of an estimator is extended to the risk of a prediction rule using statistical decision theory, and, furthermore, this framework is used to evaluate a prediction rule derived by minimizing the Bayes prediction risk, analogously to a decision rule derived by minimizing the Bayes risk. It is shown that such a rule is admissible and can be derived by minimizing the (Bayes) posterior predictive risk. These results may motivate the use of a (Bayesian) posterior predictive distribution to make a prediction, since admissibility is a desirable property, and minimizing the Bayes posterior predictive risk is computationally tractable in many cases, just as minimizing the Bayes risk is computationally straightforward under commonly used loss functions. For instance, the posterior predictive mean is admissible and minimizes the Bayesian prediction risk under a squared error loss, just as the posterior mean is admissible and minimizes the Bayesian risk under a squared error loss (under weak conditions).

\section{Definitions and Assumptions}
Denote the parameter space $\Theta \subseteq \mathbb{R}^k$. Without essential loss of generality, it is assumed all random variables are continuous and have Lebesgue-integrable density functions; the discrete case can be considered using the counting measure. Additionally, it is assumed that conditions hold to apply the Fubini theorem (e.g., Theorem 2.8 of [Lehmann and Casella]), and the prior distribution is proper, meaning it integrates to unity and is a valid probability density. Note that the inferential versions of the definitions given below are covered by many sources, including [Berger], [Robert], and [Wasserman].

\paragraph{Definition(s) 1:} Denote  the random variable $Y_{obs} \in \mathbb{R}^M$ for data observed, the random variable $Y_{pred} \in \mathbb{R}^N$ for data one would like to predict, and $\theta$ for an unknown that indexes a data generating process, $f(y_{pred},y_{obs}| \theta)$, a probability density function $f: \mathbb{R}^N \times \mathbb{R}^M \rightarrow [0,\infty)$. In the Bayesian context, one assumes that $\theta \in \mathbb{R}^k$ is a random variable and has a prior probability distribution $g(\theta)$, $g: \mathbb{R}^k \rightarrow [0,\infty)$ ascribed to it, such that the joint distribution of $(Y_{pred}, Y_{obs}, \theta)$ is given by $f(y_{pred},y_{obs}|\theta)g(\theta)$. It is further assumed that  $g(\theta) > 0$ $\forall \theta \in  \Theta$. The \textit{posterior predictive distribution} is the conditional distribution of $Y_{pred}$ given $y_{obs}$, $p(y_{pred}|y_{obs})$, $p: \mathbb{R}^N \times \mathbb{R}^M \rightarrow [0,\infty)$.
\paragraph{Definition 2:} A \textit{prediction rule} $\hat{Y}(.)$ is a function of $Y_{obs}$ meant to make a guess at $Y_{pred}$, $\hat{Y}: \mathbb{R}^M \rightarrow \mathbb{R}^N$. This is analogous to a decision rule $\hat{\theta}$ that is meant to be a guess of an unknown $\theta$. 
\paragraph{Definition 3:} A \textit{loss function} $L: \mathbb{R}^N \times \mathbb{R}^N \rightarrow [0,\infty)$, $L(\hat{Y}(Y_{obs}), Y_{pred})$, penalizes a discrepancy between $\hat{Y}(Y_{obs})$ and $Y_{pred}$. The frequentist notion of \textit{prediction risk} of the prediction rule can be considered the average of $L(\hat{Y},Y_{pred})$ holding $\theta$ fixed, just as frequentist risk of an estimator is the average of $L(\hat{\theta},\theta)$ holding $\theta$ fixed. The \textit{Bayesian prediction risk} is the average of $L(\hat{Y},Y_{pred})$ over the joint distribution of $(Y_{pred}$, $Y_{obs}, \theta$), just as the Bayes risk is the average of $L(\hat{\theta},\theta)$ over the joint distribution of $(Y_{obs}, \theta)$. Throughout, it is assumed frequentist and Bayesian prediction risk are continuous and well defined $\forall \theta \in \Theta$, $\hat{Y}$ and $Y_{pred}$.
\paragraph{Definition 4:} A \textit{Bayes prediction rule} is one which minimizes the \textit{Bayesian prediction risk}.
\paragraph{Definition 5:} A prediction rule $\hat{Y}$ is \textit{admissible} if there is no other prediction rule that achieves a frequentist prediction risk that is both: i) at least as small as that of $\hat{Y}$ $\forall \theta \in \Theta$ and ii) strictly smaller than that of $\hat{Y}$, for at least a single value of $\theta \in \Theta$. 

\section{Theorems}
\paragraph{Theorem 1:} Under the assumptions of section 2, a Bayes prediction rule can be found by minimizing the average of $L(\hat{Y}(Y_{obs}), Y_{pred})$ over $p(y_{pred}|y_{obs})$, referred to as the \textit{posterior predictive risk}. 
\paragraph{Proof:} The Bayes prediction risk is the average of $L(\hat{Y}(Y_{obs}), Y_{pred})$ over the joint distribution of $(Y_{obs},Y_{pred})$, since $L(\hat{Y}(Y_{obs}), Y_{pred})$ is not a function of $\theta$, and so $\theta$ can be integrated (or summed) out of the joint distribution of $(Y_{obs}, Y_{pred}, \theta)$. Furthermore, the distribution of $(Y_{obs},Y_{pred})$  can be written as $p(y_{pred}|y_{obs})p(y_{obs})$ (overloading the $p(.)$ notation for less clutter), so to minimize the Bayes prediction risk it suffices to minimize the average of $L(\hat{Y}(Y_{obs}), Y_{pred})$ over $p(y_{pred}|y_{obs})$ for arbitrary $y_{obs}$. In other words, the standard argument for deriving a Bayes rule can be applied by integrating out $\theta$ as a nuisance parameter.

To make this argument more explicit, mathematical notation is introduced, overloading the $p(.)$ notation to indicate probability densities. A Bayesian prediction rule is derived by minimizing the Bayesian prediction risk:
\begin{eqnarray*}
\int \int \int L(\hat{y}(y_{obs}),y_{pred})p(y_{pred},y_{obs},\theta) d\theta dy_{pred} dy_{obs} &=& \int \int L(\hat{y}(y_{obs}),y_{pred}) \int p(y_{pred},y_{obs},\theta) d\theta dy_{pred} dy_{obs} \\
                           														   &=& \int \int L(\hat{y}(y_{obs}),y_{pred}) p(y_{pred},y_{obs}) dy_{pred} dy_{obs} \\
																	   &=& \int \int L(\hat{y}(y_{obs}),y_{pred}) p(y_{pred}|y_{obs})p(y_{obs}) dy_{pred} dy_{obs} \\
																	   &=& \int \int L(\hat{y}(y_{obs}),y_{pred}) p(y_{pred}|y_{obs})dy_{pred} p(y_{obs}) dy_{obs} \
																	   \end{eqnarray*}
Which can be minimized by minimizing $\int L(\hat{y}(y_{obs}),y_{pred}) p(y_{pred}|y_{obs})dy_{pred}$ for arbitrary $y_{obs}$, i.e., the posterior predictive risk.
																	   
\paragraph{Theorem 2:} Under the assumptions of section 2, Bayesian prediction rules are admissible. 
\paragraph{Proof:} Assume there is a Bayesian prediction rule $\hat{Y}_{Bayes}$ that is not admissible, so that there exists another prediction rule $\hat{Y}_{new}$ which achieves a strictly smaller frequentist prediction risk for some value $\theta* \in \Theta$, and at least as small frequentist prediction risk for all other values in $\Theta$. Since frequentist risk is assumed to be continuous,  there is some measurable subset of $S$ of $\mathbb{R}^k$ containing $\theta*$ such that the frequentist risk of $\hat{Y}_{new}$ is strictly smaller than that of  $\hat{Y}_{Bayes}$ on this subset. In other words $\forall \theta \in S$:
\begin{eqnarray*}
\int \int L(\hat{y}_{new},y_{pred}) f(y_{pred},y_{obs}|\theta)dy_{pred}dy_{obs}  < \int \int L(\hat{y}_{Bayes},y_{pred}) f(y_{pred},y_{obs}|\theta)dy_{pred}dy_{obs}\end{eqnarray*}
and $\forall \theta \in S^c$: 
\begin{eqnarray*}
\int \int L(\hat{y}_{new},y_{pred}) f(y_{pred},y_{obs}|\theta)dy_{pred}dy_{obs}    \leq \int \int L(\hat{y}_{Bayes},y_{pred}) f(y_{pred},y_{obs}|\theta)dy_{pred}dy_{obs}  
\end{eqnarray*}
Since the Bayesian prediction risk is the average of the frequentist risk over $\theta$, and the prior $g(\theta)$ is strictly non-zero for all points in $\Theta$ (including those points for which the frequentist risk of $\hat{Y}_{new}$ is strictly smaller than that of $\hat{Y}_{Bayes}$), this yields a prediction rule with strictly smaller Bayesian prediction risk, which is a contradiction. In other words, the previous inequalities imply:
\begin{eqnarray*}
\int_S \int \int L(\hat{y}_{new},y_{pred}) f(y_{pred},y_{obs}|\theta)g(\theta)dy_{pred}dy_{obs}d\theta  < \int_S \int \int L(\hat{y}_{Bayes},y_{pred}) f(y_{pred},y_{obs}|\theta)g(\theta)dy_{pred}dy_{obs}d\theta
\end{eqnarray*} 
and:
\begin{eqnarray*}
\int_{S^c} \int \int L(\hat{y}_{new},y_{pred}) f(y_{pred},y_{obs}|\theta)g(\theta)dy_{pred}dy_{obs}d\theta  \leq \int_{S^c} \int \int L(\hat{y}_{Bayes},y_{pred}) f(y_{pred},y_{obs}|\theta)g(\theta)dy_{pred}dy_{obs}d\theta
\end{eqnarray*}  
so:
\begin{eqnarray*}
\int \int \int L(\hat{y}_{new},y_{pred}) f(y_{pred},y_{obs}|\theta)g(\theta)dy_{pred}dy_{obs}d\theta  < \int \int \int L(\hat{y}_{Bayes},y_{pred}) f(y_{pred},y_{obs}|\theta)g(\theta)dy_{pred}dy_{obs}d\theta
\end{eqnarray*}  
which contradicts that $\hat{Y}_{Bayes}$ is a Bayes rule.  
\section{Discussion}
The contribution of this note has been to show that two results of statistical decision theory can be extended to a prediction setting with natural modifications to standard textbook definitions. Namely, the loss function allows both arguments to be random as opposed to one fixed and one random, and the underlying ``true" state of nature $\theta$ is treated as a nuisance parameter when deriving a Bayes rule.

\end{document}